\documentclass[letterpaper]{article} 
\usepackage{aaai25}  
\usepackage{times}  
\usepackage{helvet}  
\usepackage{courier}  
\usepackage[hyphens]{url}  
\usepackage{graphicx} 
\usepackage{natbib}  
\usepackage{caption} 
\usepackage{algorithm}
\usepackage{algorithmic}

\usepackage{amsmath, bm}
\usepackage{mathtools}
\usepackage{multirow}
\usepackage{booktabs}
\usepackage{tabularx}

\usepackage{newfloat}
\usepackage{listings}
\DeclareCaptionStyle{ruled}{labelfont=normalfont,labelsep=colon,strut=off} 
\floatstyle{ruled}
\newfloat{listing}{tb}{lst}{}
\floatname{listing}{Listing}
\title{Few-Shot, No Problem: Descriptive Continual Relation Extraction}

\author{
    Nguyen Xuan Thanh \textsuperscript{\rm 1}\equalcontrib, 
    Anh Duc Le\textsuperscript{\rm 2}\equalcontrib, Quyen Tran \textsuperscript{\rm 3}\equalcontrib, Thanh-Thien Le \textsuperscript{\rm 3}\equalcontrib, \\
    Linh Ngo Van \textsuperscript{\rm 2}\thanks{Corresponding Author}, Thien Huu Nguyen \textsuperscript{\rm 4}
}
\affiliations{
     \textsuperscript{\rm 1}Oraichain Labs,  \\
    \textsuperscript{\rm 2}Hanoi University of Science and Technology,  \\
     \textsuperscript{\rm 3}VinAI Research, \\
    \textsuperscript{\rm 4}University of Oregon\\
    thanh.nx@orai.io\\
    anh.ld204628@sis.hust.edu.vn\\
   \{v.quyentt15,v.thienlt3\}@vinai.io, \\
     linhnv@soict.hust.edu.vn, thien@cs.uoregon.edu
}

\usepackage{bibentry}

\begin{document}

\maketitle

\begin{abstract}
Few-shot Continual Relation Extraction is a crucial challenge for enabling AI systems to identify and adapt to evolving relationships in dynamic real-world domains. Traditional memory-based approaches often overfit to limited samples, failing to reinforce old knowledge, with the scarcity of data in few-shot scenarios further exacerbating these issues by hindering effective data augmentation in the latent space. In this paper, we propose a novel retrieval-based solution, starting with a large language model to generate descriptions for each relation. From these descriptions, we introduce a bi-encoder retrieval training paradigm to enrich both sample and class representation learning. Leveraging these enhanced representations, we design a retrieval-based prediction method where each sample "retrieves" the best fitting relation via a reciprocal rank fusion score that integrates both relation description vectors and class prototypes. Extensive experiments on multiple datasets demonstrate that our method significantly advances the state-of-the-art by maintaining robust performance across sequential tasks, effectively addressing catastrophic forgetting.
\end{abstract}

\section{Introduction}
Relation Extraction (RE) refers to classifying semantic relationships between entities within text into predefined types.  Conventional RE tasks assume all relations are present at once, ignoring the fact that new relations continually emerge in the real world.  Few-shot Continual Relation Extraction (FCRE) is a subfield of continual learning \cite{hai2024continual,linhp,phan2022reducing, tran2024leveraginghierarchicaltaxonomiespromptbased, tran2024koppaimprovingpromptbasedcontinual,le2024mixture} where a model must continually assimilate new emerging relations while avoiding the forgetting of old ones, a task made even more challenging by the limited training data available. The importance of FCRE stems from its relevance to dynamic real-world applications, garnering increasing interest in the field \cite{ chen-etal-2023-consistent, le2024continual,le2025aaai}.

\begin{figure}
    \centering
    \includegraphics[width=\linewidth]{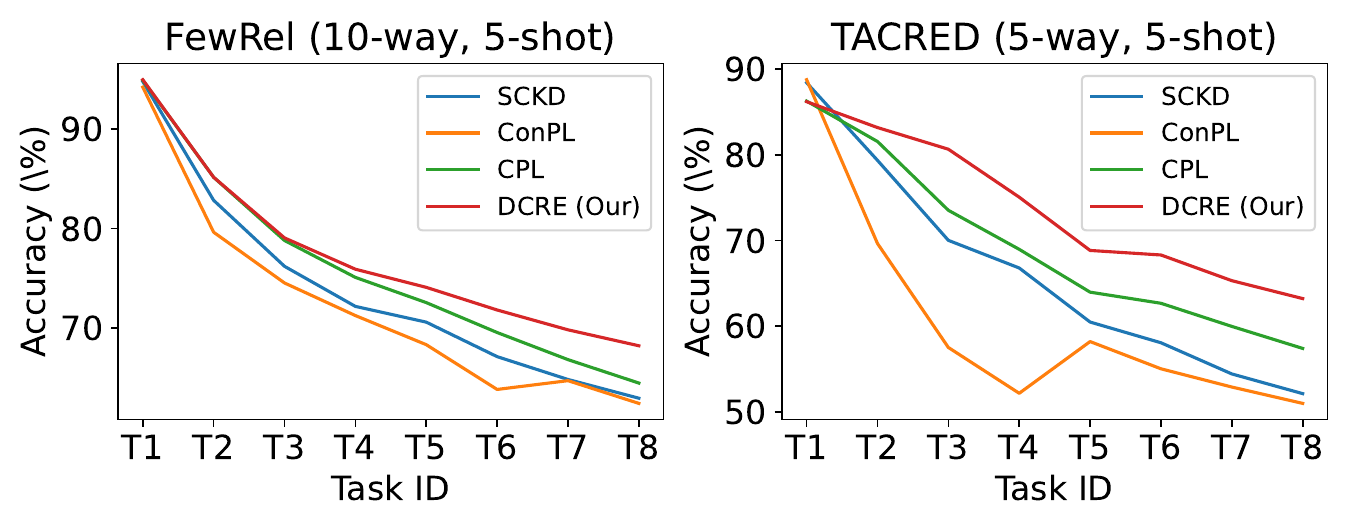}
    \caption{Existing FCRE methods face catastrophic forgetting due to the limited and poor quality of old training samples stored in the memory buffer.}
    \label{fig:line_plot_bert}
\end{figure}

State-of-the-art approaches to FCRE often rely on memory-based methods for continual learning  \cite{lopez2017gradient, nguyen2023spectral, le2024sharpseq,dao2024lifelong}. However, these methods frequently suffer from overfitting to the limited samples stored in memory buffers. This overfitting hampers the reinforcement of previously learned knowledge, leading to catastrophic forgetting—a marked decline in performance on learnt relations when new ones are introduced (Figure \ref{fig:line_plot_bert}). The few-shot scenario of FCRE exacerbates these issues, as the scarcity of data not only impedes learning on new tasks, but also hinders helpful data augmentation, which are crucial in many methods \cite{shin2017continual}.

In order to improve on these methods, we must not completely disregard them or dwell on their weaknesses, but rather contemplate their biggest strength. \emph{Why do so many methods use the memory buffer in the first place?} The primary objective of these replay buffers is to rehearse and reinforce past knowledge, providing the model with something to "look back" at during training. However, these past samples may not always be representative of the entire class and can still lead to sub-optimal performance. Based on this observation, we propose a straightforward: besides relying on potentially unrepresentative past samples, we leverage our knowledge of the past relations themselves. This insight leads to our approach of generating detailed descriptions for each relation. These descriptions inherently represent the class more accurately than the underlying information from a set of samples, serving as stable pivots for the model to align with past knowledge while learning new information. By using these descriptions, we create a more robust and effective method for Few-Shot Continual Relation Extraction, ensuring better retention of knowledge across tasks.

Overall, our paper makes the following contributions:
\begin{description}
    \item[a.] We introduce an innovative approach to Few-Shot Continual Relation Extraction that leverages Large Language Models (LLMs) to generate comprehensive descriptions for each relation. These descriptions serve as stable class representations in the latent space during training. Unlike the variability and limitations of a limited set of samples from the memory buffer, these descriptions define the inherent meaning of the relations, offer a more reliable anchor, significantly reducing the risk of catastrophic forgetting. Importantly, LLMs are employed exclusively for generating descriptions and do not participate in the training or inference processes, ensuring that our method incurs minimal computational overhead.

    \item[b.] We design a bi-encoder retrieval learning framework for both sample and class representation learning. In addition to sample representation contrastive learning, we integrate a description-pivot learning process, ensuring alignment of samples which maximize their respective class samples proximity, while non-matching samples are distanced.
    
    \item[c.] Building on the enhanced representations, we introduce the \emph{Descriptive Retrieval Inference} (DRI) strategy. In this approach, each sample "retrieves" the most fitting relation using a reciprocal rank fusion score that integrates both class descriptions and class prototypes, effectively finalizing the retrieval-based paradigm that underpins our method.
\end{description}


\section{Background}
\subsection{Problem Formulation}
\label{sec:bg:problem}

In Few-Shot Continual Relation Extraction (FCRE), a model must continuously assimilate new knowledge from a sequential series of tasks. For each $t$-th task, the model undergoes training on the dataset $D^t = \{ (x^t_i, y^t_i)\}_{i=1}^{N \times K}$. Here, $N$ represents the number of relations in the task $R^t$, and $K$ denotes the limited number of samples per relation, reflecting the few-shot learning scenario. Each sample $(x, y)$ includes a sentence $x$ containing a pair of entities $(e_h, e_t)$ and a relation label $y \in R$. This type of task setup is referred to as \textit{"N-way-K-shot"} \cite{chen-etal-2023-consistent}. Upon completion of task $t$, the dataset $D^t$ should not be extensively included in subsequent learning, as continual learning aims to avoid retraining on all prior data. Ultimately, the model's performance is assessed on a test set which encompasses all encountered relations $\tilde{R}^T = \bigcup_{t=1}^T R^t$.

For clarity, each task in FCRE can be viewed as a conventional relation extraction problem, with the key challenge being the scarcity of samples available for learning. The primary goal of FCRE is to develop a model that can consistently acquire new knowledge from limited data while retaining competence in previously learned tasks. In the following subsections, we will explore the key aspects of FCRE models as addressed by state-of-the-art studies.

\subsection{Encoding Latent Representation}
\label{sec:bg:rep_encode}




A key initial consideration in Relation Extraction is how to \emph{formalize the latent representation} of the input, as the output of a Transformer \cite{vaswani2017attention} is a matrix. In this work, we adopt a method recently introduced by \citet{ma-etal-2024-making}. Given an input sentence $x$, which includes a head entity $e_h$ and a tail entity $e_t$, we reformulate it into a Cloze-style phrase $T(x)$ by incorporating a \texttt{[MASK]} token, which represents the relation between the entities. Specifically, the template is structured as follows:
\begin{align}
\begin{aligned}
  T({x}) = \; &x \left[v_{0:n_0-1}\right] e_h \left[v_{n_0:n_1-1}\right] [\texttt{MASK}] \\
  &\left[v_{n_1:n_2-1}\right] e_t \left[v_{n_2:n_3-1}\right].
\label{eq:template}
\end{aligned}
\end{align}
 Each $[v_i]$ denotes a learnable continuous token, and $n_j$ determines the number of tokens in each phrase. In our specific implementation, we use BERT's \texttt{[UNUSED]} tokens as $[v]$. The soft prompt phrase length is set to $3$ tokens, meaning $n_0$, $n_1$, $n_2$ and $n_3$ correspond to the values of $3$, $6$, $9$, and $12$, respectively. We then forward the templated sentence $T({x})$ through BERT to encode it into a sequence of continuous vectors, from which we obtain the hidden representation $\bm{z}$ of the input, corresponding to the position of the \texttt{[MASK]} token:
\begin{equation}
    \bm{z} = [\mathcal{M} \circ T] ({x})[\rm{position}(\texttt{[MASK]})],
\end{equation}
where $\mathcal{M}$ denotes the backbone pre-trained language model. This latent representation is then passed through an MLP for prediction, enabling the model to learn which relation that best fills the \texttt{[MASK]} token. 



\subsection{Learning Latent Representation}
\label{sec:bg:rep_learning}

In conventional Relation Extraction scenarios, a basic framework typically employs a backbone PLM followed by an MLP classifier to directly map the input space to the label space using Cross Entropy Loss. However, this approach proves inadequate in data-scarce settings \cite{snell2017prototypical}. Consequently, training paradigms which directly target the latent space, such as contrastive learning, emerge as more suitable approaches. To enhance the semantic richness of the information extracted from the training samples, two popular losses are often utilized: \emph{Supervised Contrastive Loss} and \emph{Hard Soft Margin Triplet Loss}.

\paragraph{Supervised Contrastive Loss.}
To enhance the model's discriminative capability, we employ the Supervised Contrastive Loss (SCL) \cite{khosla2020supcon}. This loss function is designed to bring positive pairs of samples, which share the same class label, closer together in the latent space. Simultaneously, it pushes negative pairs, belonging to different classes, further apart. Let $\bm{z}_x$ represent the hidden vector output of sample ${x}$, the positive pairs $(\bm{z}_x, \bm{z}_p)$ are those who share a class, while the negative pairs $(\bm{z}_x, \bm{z}_n)$ correspond to different labels. The SCL is computed as follows:
\begin{gather}
\mathcal{L}_{\textrm{SC}}(x) = - \sum_{{p} \in P({x})} \log \frac{
    f(\bm{z}_x, \bm{z}_p)
}
{\sum_{{u} \in \mathcal{D}\setminus \{x\}} f(\bm{z}_x, \bm{z}_u)} 
\end{gather}
where $f(\mathbf{x}, \mathbf{y}) \coloneqq \exp\left(\frac{\gamma(\mathbf{x}, \mathbf{y})}{\tau}\right)$, $\gamma(\cdot, \cdot)$ denotes the cosine similarity function, and $\tau$ is the temperature scaling hyperparameter. $P({x})$ and $\mathcal{D}$ denote the sets of positive samples with respect to sample ${x}$ and the training set, respectively.

\paragraph{Hard Soft Margin Triplet Loss.}
To achieve a balance between flexibility and discrimination, the Hard Soft Margin Triplet Loss (HSMT) integrates both hard and soft margin triplet loss concepts \cite{hermans2017defense}. This loss function is designed to maximize the separation between the most challenging positive and negative samples, while preserving a soft margin for improved flexibility. Formally, the loss is defined as:
\begin{multline}
\mathcal{L}_{\textrm{ST}}({x}) = \\
- \log \bigg(1 \ + \max_{{p} \in P(\bm{x})} e^{\xi(\bm{z}_x, \bm{z}_p)}
- \min_{{n} \in N({x})} e^{ \xi(\bm{z}_x, \bm{z}_n)}
\bigg),
\end{multline}
where $\xi(\cdot, \cdot)$ denotes the Euclidean distance function. The objective of this loss is to ensure that the hardest positive sample is as distant as possible from the hardest negative sample, thereby enforcing a flexible yet effective margin.

During training, these two losses is aggregated and referred to as the \emph{Sample-based learning loss}:
\begin{equation}
\mathcal{L}_\text{Samp} = \beta_\text{SC} \cdot \mathcal{L}_\text{SC} + \beta_\text{ST} \cdot \mathcal{L}_\text{ST}
\end{equation}

\section{Proposed Method}

\subsection{Label Descriptions}
\label{sec:method:descriptions}

\begin{figure}
    \centering
    \includegraphics[width=\linewidth]{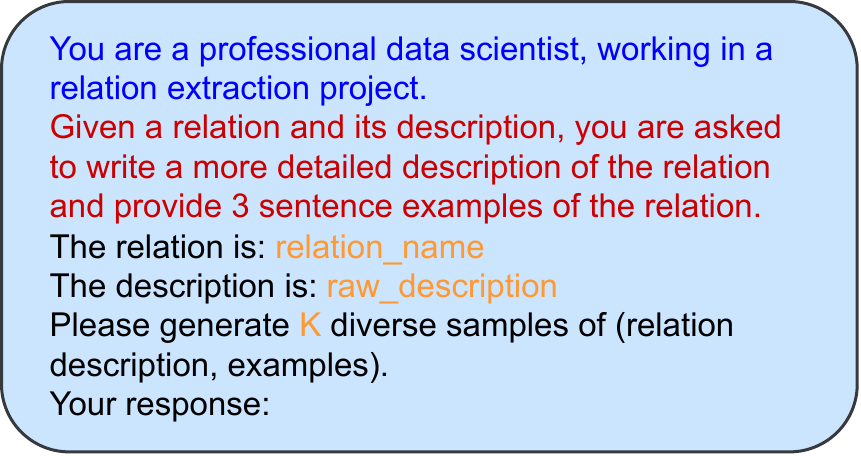}
    \caption{Prompt to generate relation descriptions with LLMs.}
    \label{fig:prompt}
\end{figure}

A core component of our method is achieving robust class latent representations, making class encoding crucial. To this end, having detailed definitions for each label, alongisde the hidden information extracted from the samples, is essential for our approach. In fact, the datasets used for benchmarking already provide each relation with a concise description, which we refer to as the \emph{Raw description}. While leveraging these descriptions has shown promise in previous work \cite{luo2024synergistic}, this approach remains limited due to its reliance on a one-to-one mapping between input embeddings and a single label description representation per task. This singular approach fails to offer rich, diverse, and robust information about the labels, leading to potential noise, instability, and suboptimal model performance.

To address these limitations, we employ Gemini 1.5 \cite{team2023gemini, reid2024gemini} to generate $K$ diverse, detailed, and illustrative descriptions for each relation. In particular, for each label, the respective raw description will be fed into the \emph{LLM prompt}, serving as an expert-in-the-loop to guide the model. Our prompt template is depicted in Figure \ref{fig:prompt}.

\subsection{Description-pivot Learning}
\label{sec:method:retrieval}

The single most valuable quality of class descriptions in our problem is that they are literal definitions of a relation, which makes them more accurate representations of that class than the underlying information from a set of samples. Thanks to this strength, they serve as stable knowledge anchors for the model to rehearse from, enabling effective reinforcement of old knowledge while assimilating new information. Unlike the variability of individual samples, a description remains consistent, providing a more reliable reference point for the model to rehearse from, effectively mitigating catastrophic forgetting.

To fully leverage this inherent advantage, we integrate these descriptions into the training process, framing the task as one of retrieving definition, which embodies real-world meaning, rather than a straightforward categorical classification. By doing so, we capitalize on the unchanging nature of descriptions, making them the focal point of our model's learning. Specifically, we incorporate two description-centric losses to enhance this retrieval-oriented approach:
\begin{equation}
    \mathcal{L}_\text{Des} = \beta_\text{HM}\cdot \mathcal{L}_\text{HM} + \beta_\text{MI}\cdot\mathcal{L}_\text{MI}.
\end{equation}
Here, $\mathcal{L}_\text{HM}$ and $\mathcal{L}_\text{MI}$ denote the Hard Margin Loss and the Mutual Information Loss, respectively. These losses are elaborated upon in the following paragraphs.

\paragraph{Hard Margin Loss.}
The Hard Margin Loss leverages label descriptions to refine the model's ability to distinguish between hard positive and hard negative pairs. Given the output hidden vectors $\{ \bm{d}^k_x \}_{k=1, ..., K}$ from BERT corresponding to the label description of sample ${x}$, and $\bm{z}_p$ and $\bm{z}_n$ representing the hidden vectors of positive and negative samples respectively, the loss function is formulated to maximize the alignment between $ \bm{d}^k_x $ and its corresponding positive sample, while enforcing a strict margin against negative samples. Specifically, the loss is formulated as follows:
\begin{align}
\mathcal{L}_{\textrm{HM}}(x) &= \sum_{k=1}^K \mathcal{L}_{\textrm{HM}}^k(x), \\
\mathcal{L}_{\textrm{HM}}^k(x) &= 
\sum_{p \in P_\textrm{H}(x)} (1 - \gamma(\bm{d}^k_x, \bm{z}_p))^2 \notag \\
&+ \sum_{n \in N_\textrm{H}(x)} max(0, m - 1 + \gamma(\bm{d}^k_x, \bm{z}_n) )^2,
\end{align}
where $m$ is a margin hyperparameter; $\gamma(\cdot, \cdot)$ denotes the cosine similarity function; $P_\textrm{H}(x)$ and $N_\textrm{H}(x)$ represent the sets of hard positive and hard negative samples, respectively. They are determined by comparing the similarity between $\bm{d}^k_x$ and both positive and negative pairs, specifically focusing on the most challenging pairs where the similarity to negative samples is close to or greater than that of positive samples, defined as follows:
\begin{align}
\begin{aligned}
    P_\textrm{H}(x) = \{&p \in P(x) | 1 - \gamma(\bm{d}^k_x, \bm{z}_p) \\
    &> min_{n \in N(x)}(1 - \gamma(\bm{d}^k_x, \bm{z}_n)), \forall k \in [K] \},
\end{aligned} \\
\begin{aligned}
    N_\textrm{H}(x) = \{&n \in N(x) | 1 - \gamma(\bm{d}^k_x, \bm{z}_n) \\
    &< max_{p \in P(x)}(1 - \gamma(\bm{d}^k_x, \bm{z}_p)), \forall k \in [K] \}.
\end{aligned}
\end{align}


By utilizing the label description vectors $\{ \bm{d}^k_x \}$, optimizing $\mathcal{L}_{\textrm{HM}}(x)$ effectively sharpens the model's decision boundary, reducing the risk of confusion between similar classes and improving overall performance in few-shot learning scenarios. The loss penalizes the model more heavily for misclassifications involving these hard samples, ensuring that the model pays particular attention to the most difficult cases, thereby enhancing its discriminative power.

\begin{figure}[t]
    \centering
    \includegraphics[width=\linewidth]{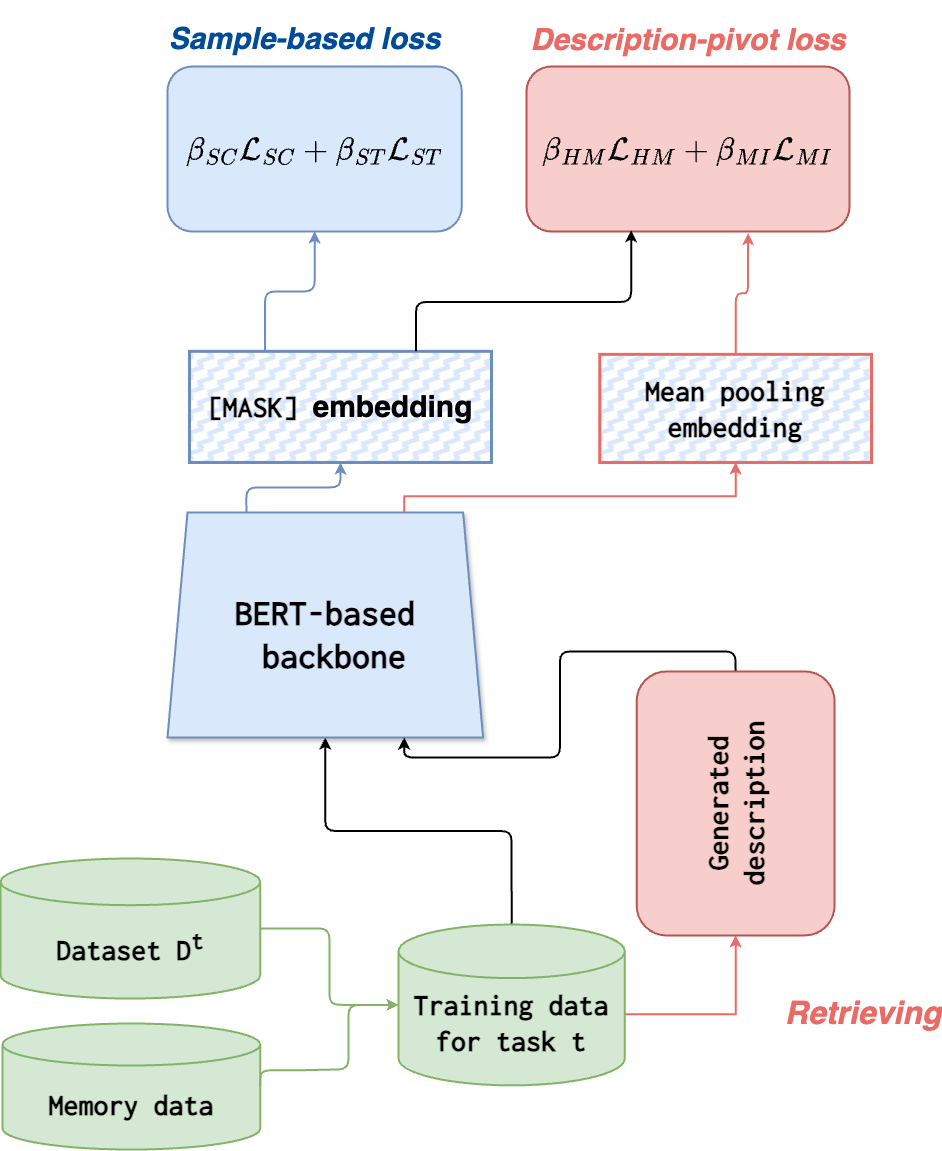}
    \caption{Our Framework.}
    \label{fig:framework}
\end{figure}

\paragraph{Mutual Information Loss.}
The Mutual Information (MI) Loss is designed to maximize the mutual information between the input sample's hidden representation $\bm{z_x}$ of $\bm{x}$ and its corresponding retrieved descriptions, promoting a more informative alignment between them. Let $\bm{d}_n$ be a hidden vector of other label descriptions than $\bm{x}$. According to \citet{DBLP:journals/corr/abs-1807-03748}, the Mutual Information $MI(x)$ between the input embedding $\bm{z}_x$ and its corresponding label description follows the following inequation:
\begin{equation}
    MI \geq \log B + \textnormal{InfoNCE} (\{ x_i\}_{i=1}^B; h),
\end{equation}
where we have defined:
\begin{multline}
\textnormal{InfoNCE} (\{ x_i\}_{i=1}^B; h) = \\
\frac{1}{B}\sum_{i=1}^B \log \frac{\sum_{k=1}^K h(\bm{z_i}, \bm{d}_i^k)}{\sum_{j=1}^B \sum_{k=1}^K h(\bm{z_j}, \bm{d}_j^k)},
\end{multline}
where $h(\bm{z}_j, \bm{d}_j^k) = \exp \left(\frac{\bm{z}_j^TW\bm{d}_j^k}{\tau}\right)$. Here, $\tau$ is the temperature, $B$ is mini-batch size and $W$ is a trainable parameter. Finally, the MI loss function in our implementation is: 
\begin{multline}
\mathcal{L}_{\textrm{MI}}(x) = \\
- \log \frac{ \sum_{k=1}^K
    h(\bm{z}_x, \bm{d}_x^k)
}{
    \sum_{k=1}^K
    h(\bm{z}_x, \bm{d}_x^k)
    + \sum_{n \in N(x)} \sum_{k=1}^K h(\bm{z}_x, \bm{d}_n^k)
}
\end{multline}

This loss ensures that the representation of the input sample is strongly associated with its corresponding label, while reducing its association with incorrect labels, thereby enhancing the discriminative power of the model.

\begin{table*}[!ht]
    \centering
    \setlength{\tabcolsep}{1mm}
    \begin{tabular}{lllllllll|c}
        \multicolumn{3}{l}{\textbf{FewRel} \textit{(10-way--5-shot)}} \\
        \hline
        Method & $\mathcal{T}^1$ & $\mathcal{T}^2$ & $\mathcal{T}^3$ & $\mathcal{T}^4$ & $\mathcal{T}^5$ & $\mathcal{T}^6$ & $\mathcal{T}^7$ & $\mathcal{T}^8$ & $\Delta \downarrow$ \\ 
        \hline \hline
        RP-CRE      & $93.97_{\pm 0.64}$ & $76.05_{\pm 2.36}$ & $71.36_{\pm 2.83}$ & $69.32_{\pm 3.98}$ & $64.95_{\pm 3.09}$ & $61.99_{\pm 2.09}$ & $60.59_{\pm 1.87}$ & $59.57_{\pm 1.13}$ & 34.40 \\
        CRL         & $94.68_{\pm 0.33}$ & $80.73_{\pm 2.91}$ & $73.82_{\pm 2.77}$ & $70.26_{\pm 3.18}$ & $66.62_{\pm 2.74}$ & $63.28_{\pm 2.49}$ & $60.96_{\pm 2.63}$ & $59.27_{\pm 1.32}$ & 35.41\\
        CRECL       & $93.93_{\pm 0.22}$ & $82.55_{\pm 6.95}$ & $74.13_{\pm 3.59}$ & $69.33_{\pm  3.87}$ & $66.51_{\pm 4.05}$ & $64.60_{\pm 1.92}$ & $62.97_{\pm 1.46}$ & $59.99_{\pm 0.65}$ & 33.94\\
        ERDA        & $92.43_{\pm 0.32}$ & $64.52_{\pm 2.11}$ & $50.31_{\pm 3.32}$ & $44.92_{\pm 3.77}$ & $39.75_{\pm 3.34}$ & $36.36_{\pm 3.12}$ & $34.34_{\pm 1.83}$ & $31.96_{\pm 1.91}$ & 60.47\\
        SCKD        & $94.77_{\pm 0.35}$ & $82.83_{\pm 2.61}$ & $76.21_{\pm 1.61}$ & $72.19_{\pm 1.33}$ & $70.61_{\pm 2.24}$ & $67.15_{\pm 1.96}$ & $64.86_{\pm 1.35}$ & $62.98_{\pm 0.88}$ & 31.79\\ 
        ConPL$^{**}$ & $\mathbf{95.18_{\pm 0.73}}$ & $79.63_{\pm 1.27}$ & $74.54_{\pm 1.13}$ & $71.27_{\pm 0.85}$ & $68.35_{\pm 0.86}$ & $63.86_{\pm 2.03}$ & $64.74_{\pm 1.39}$ & $62.46_{\pm 1.54}$ &32.72 \\
        CPL & {94.87} & {85.14} & {78.80} & {75.10} & {72.57} & {69.57} & {66.85} & {64.50} & {30.37}\\ 
        CPL + MI  & $94.69_{\pm 0.7}$ & $\mathbf{85.58_{\pm 1.88}}$ & $\mathbf{80.12_{\pm 2.45}}$ & \underline{${75.71_{\pm 2.28}}$} & \underline{${73.90_{\pm 1.8}}$} & \underline{${70.72_{\pm 0.91}}$} & \underline{${68.42_{\pm 1.77}}$} & \underline{${66.27_{\pm 1.58}}$} & 28.42\\
        DCRE & \underline{$94.93_{\pm 0.39}$} & \underline{$85.14_{\pm 2.27}$} & \underline{${79.06_{\pm 1.68}}$} & $\mathbf{75.92_{\pm 2.03}}$ & $\mathbf{74.10_{\pm 2.53}}$ & $\mathbf{71.83_{\pm 2.17}}$ & $\mathbf{69.84_{\pm 1.48}}$ & $\mathbf{68.24_{\pm 0.79}}$ & \textbf{26.69} \\
        \hline
        \\
         
        \multicolumn{3}{l}{\textbf{TACRED} \textit{(5-way-5-shot)}} \\
    \hline
        Method & $\mathcal{T}^1$ & $\mathcal{T}^2$ & $\mathcal{T}^3$ & $\mathcal{T}^4$ & $\mathcal{T}^5$ & $\mathcal{T}^6$ & $\mathcal{T}^7$ & $\mathcal{T}^8$ & $\Delta \downarrow$ \\ \hline \hline
        RP-CRE      & $87.32_{\pm 1.76}$ & $74.90_{\pm 6.13}$ & $67.88_{\pm 4.31}$ & $60.02_{\pm 5.37}$ & $53.26_{\pm 4.67}$ & $50.72_{\pm 7.62}$ & $46.21_{\pm 5.29}$ & $44.48_{\pm 3.74}$ & 42.84\\
        CRL         & $88.32_{\pm 1.26}$ & $76.30_{\pm 7.48}$ & $69.76_{\pm 5.89}$ & $61.93_{\pm 2.55}$ & $54.68_{\pm 3.12}$ & $50.92_{\pm 4.45}$ & $47.00_{\pm 3.78}$ & $44.27_{\pm 2.51}$ & 44.05\\
        CRECL       & $87.09_{\pm 2.50}$ & $78.09_{\pm 5.74}$ & $61.93_{\pm 4.89}$ & $55.60_{\pm 5.78}$ & $53.42_{\pm 2.99}$ & $51.91_{\pm 2.95}$ & $47.55_{\pm 3.38}$ & $45.53_{\pm 1.96}$ & 41.56\\
        ERDA        & $81.88_{\pm 1.97 }$ & $53.68_{\pm 6.31}$ & $40.36_{\pm 3.35}$ & $36.17_{\pm 3.65}$ & $30.14_{\pm 3.96}$ & $22.61_{\pm 3.13}$ & $22.29_{\pm 1.32}$ & $19.42_{\pm 2.31}$ & 62.46\\
        SCKD        & \underline{$88.42_{\pm 0.83}$} & $79.35_{\pm 4.13}$ & $70.61_{\pm 3.16}$ & $66.78_{\pm 4.29}$ & $60.47_{\pm 3.05}$ & $58.05_{\pm 3.84}$ & $54.41_{\pm 3.47}$ & $52.11_{\pm 3.15}$ & 36.31\\
        ConPL$^{**}$ & {$\mathbf{88.77_{\pm 0.84}}$} & $69.64_{\pm 1.93}$ & $57.50_{\pm 2.48}$ & $52.15_{\pm 1.59}$ & $58.19_{\pm 2.31}$ & $55.01_{\pm 3.12}$ & $52.88_{\pm 3.66}$ & $50.97_{\pm 3.41}$ & 37.80\\ 
        CPL & 86.27 & \underline{81.55} & {73.52} & \underline{68.96} & {63.96} & {62.66} & {59.96} & {57.39} & {28.88} \\
        CPL + MI   & $85.67_{\pm 0.8}$ & \underline{$82.54_{\pm 2.98}$}& \underline{$75.12_{\pm 3.67}$} & \underline{$70.65_{\pm 2.75}$} & \underline{$66.79_{\pm 2.18}$} & \underline{$65.17_{\pm 2.48}$} & \underline{$61.25_{\pm 1.52}$} & \underline{$59.48_{\pm 3.53}$} & 26.19 \\ 
        DCRE & $86.20_{\pm 1.35}$ & {$\mathbf{83.18_{\pm 8.04}}$} & {$\mathbf{80.65_{\pm 3.06}}$} & {$\mathbf{75.05_{\pm 3.07}}$} & {$\mathbf{68.83_{\pm 5.05}}$} & {$\mathbf{68.30_{\pm 4.28}}$} & {$\mathbf{65.30_{\pm 2.74}}$} & {$\mathbf{63.21_{\pm 2.39}}$} & \textbf{22.99} \\

    \hline      
    \end{tabular}
    \caption{Accuracy (\%) of methods using BERT-based backbone after training for each task. The best results are in \textbf{bold}. **Results of ConPL are reproduced}
    \label{table:main}
\end{table*}
\paragraph{Joint Training Objective Function.}
Our model is trained using a combination of the \emph{Sample-based learning loss} mentioned in Section \ref{sec:bg:rep_learning} and our description-pivot loss $\mathcal{L}_\text{Des}$, weighted by their respective coefficients:
\begin{align}
\mathcal{L}(x) &= \mathcal{L}_\text{Samp} + \mathcal{L}_\text{Des} \\
&= \beta_\textrm{SC} \cdot \mathcal{L}_{\textrm{SC}}(x) + \beta_\textrm{ST} \cdot \mathcal{L}_{\textrm{ST}}(x) \notag \\
&\quad + \beta_\textrm{HM} \cdot \mathcal{L}_{\textrm{HM}}(x) + \beta_\textrm{MI} \cdot \mathcal{L}_{\textrm{MI}}(x),
\end{align}
where $\beta_\textrm{SC}$, $\beta_\textrm{ST}$, $\beta_\textrm{HM}$, and $\beta_\textrm{MI}$ are hyperparameters. This joint objective enables the model to leverage the strengths of each individual loss, facilitating robust and effective learning in Few-Shot Continual Relation Extraction tasks.

\paragraph{Training Procedure.}
Algorithm \ref{alg:Framework} outlines the end-to-end training process at each task $\mathcal{T}^j$, with $\Phi_{j-1}$ denoting the model after training on the previous $j-1$ tasks. In line with memory-based continual learning methods, we maintain a memory buffer $\tilde{M}_{j-1}$ that stores a few representative samples from all previous tasks ${\mathcal{T}^1, \dots, \mathcal{T}^{j-1}}$, along with a relation description set $\tilde{E}_{j-1}$ that holds the descriptions of all previously encountered relations.

\begin{enumerate}
    \item \textbf{Initialization} (Line 1--2): 
    The model for the current task, $\Phi_j$, is initialized with the parameters of $\Phi_{j-1}$. We update the relation description set $\tilde{E}_j$ by incorporating new relation descriptions from $E_j$.
    
    \item \textbf{Training on the Current Task} (Line 3): 
    We train $\Phi_j$ on $D_j$ to learn the novel relations introduced in in $\mathcal{T}^j$.
    
    \item \textbf{Memory Update} (Lines 4--8):
    We select $L$ representative samples from $D_j$ for each relation $r \in R_j$. These are the $L$ samples whose latent representations are closest to the $1$-means centroid of all class samples. These samples constitute the memory $M_r$, leading to an updated overall memory $\tilde{M}_j = \tilde{M}_{j-1} \cup M_j$ and an updated relation set $\tilde{R}_j = \tilde{R}_{j-1} \cup R_j$.

    \item \textbf{Prototype Storing} (Line 9):
    A prototype set $\tilde{P}_j$ is generated based on the updated memory $\tilde{M}_j$.
    We generate a prototype set $\tilde{P}_j$ based on the updated memory $\tilde{M}_j$.

    \item \textbf{Memory Training} (Line 10):
    We refine $\Phi_j$ by training on the augmented memory set $\tilde{M}_j^*$, ensuring that the model preserves knowledge of relations from previous tasks.
    

\end{enumerate}



\begin{algorithm}[ht]
\caption{Training procedure at each task $\mathcal{T}^j$}
\label{alg:Framework}
\textbf{Input}: $\Phi_{j-1}, \tilde{R}_{j-1}, \tilde{M}_{j-1},\tilde{K}_{j-1}, D_j, R_j, K_j$. \\
\textbf{Output}: $\Phi_j, \tilde{M}_j, \tilde{K}_j, \tilde{P}_j.$
\begin{algorithmic}[1]
\STATE Initialize $\Phi_j$ from $\Phi_{j-1}$ 
\STATE $\tilde{K}_j \leftarrow \tilde{K}_{j-1} \cup K_j$
\STATE Update $\Phi_j$ by L on $D_j$ (train on current task)
\STATE $\tilde{M}_j \leftarrow \tilde{M}_{j-1}$
\FOR{each $r \in R_j$} 
    \STATE pick $L$ samples in $D_j$ and add them into $\tilde{M}_j$
\ENDFOR
\STATE $\tilde{R}_j \leftarrow \tilde{R}_{j-1} \cup R_j$
\STATE Update $\tilde{P}_j$ with new data in ${D}_j$ (for inference)
\STATE Update $\Phi_j$ by $\mathcal{L}$ on $\tilde{M}_j$ and $D_j^*$ (train on memory)
\end{algorithmic}
\end{algorithm}




\begin{table*}[!ht]
    \centering
    \setlength{\tabcolsep}{1mm}
    \begin{tabular}{lllllllll|c}
        \multicolumn{3}{l}{\textbf{FewRel} \textit{(10-way--5-shot)}} \\
        \hline
        Method & $\mathcal{T}^1$ & $\mathcal{T}^2$ & $\mathcal{T}^3$ & $\mathcal{T}^4$ & $\mathcal{T}^5$ & $\mathcal{T}^6$ & $\mathcal{T}^7$ & $\mathcal{T}^8$ & $\Delta \downarrow$ \\ 
        \hline \hline
        CPL & $\mathbf{97.25_{\pm 0.30}}$ & $\mathbf{89.29_{\pm 2.51}}$ & $85.56_{\pm 1.21}$ & $82.10_{\pm 2.02}$ & $79.96_{\pm 2.72}$ & $78.41_{\pm 3.22}$ & $76.42_{\pm 2.25}$ & $75.20_{\pm 2.33}$ & 22.05 \\
        DCRE & $96.92_{\pm 0.16}$ & ${88.95_{\pm 1.72}}$ & $\mathbf{87.12_{\pm 1.52}}$ & {$\mathbf{85.44_{\pm 1.91}}$} & {$\mathbf{84.89_{\pm 2.12}}$} & {$\mathbf{83.52_{\pm 1.46}}$} & {$\mathbf{81.64_{\pm 0.69}}$} & {$\mathbf{80.34_{\pm 0.55}}$} & \textbf{16.58} \\
        \hline \\
        
        \multicolumn{3}{l}{\textbf{TACRED} \textit{(5-way-5-shot)}} \\
    \hline
        Method & $\mathcal{T}^1$ & $\mathcal{T}^2$ & $\mathcal{T}^3$ & $\mathcal{T}^4$ & $\mathcal{T}^5$ & $\mathcal{T}^6$ & $\mathcal{T}^7$ & $\mathcal{T}^8$ & $\Delta \downarrow$ \\ \hline
        \hline
        CPL & $88.74_{\pm 0.44}$ & $85.16_{\pm 5.38}$ & $78.35_{\pm 4.46}$ & $77.50_{\pm 4.04}$ & $76.01_{\pm 5.04}$ & $76.30_{\pm 4.41}$ & $74.51_{\pm 5.06}$ & $73.83_{\pm 4.91}$ & 14.91 \\
        DCRE & {$\mathbf{89.06_{\pm 0.59}}$} & {$\mathbf{87.41_{\pm 5.54}}$} & {$\mathbf{84.91_{\pm 3.38}}$} & {$\mathbf{84.18_{\pm 2.44}}$} & {$\mathbf{82.74_{\pm 3.64}}$} & {$\mathbf{81.92_{\pm 2.33}}$} & {$\mathbf{79.34_{\pm 2.89}}$} & {$\mathbf{79.10_{\pm 2.37}}$} & \textbf{9.96}  \\

    \hline
      
    \end{tabular}
    \caption{Accuracy (\%) of methods using LLM2Vec-based backbone after training for each task. The best results are in \textbf{bold}.}
    \label{table:main_llmvec}
\end{table*}

\subsection{Descriptive Retrieval Inference}
\label{sec:method:relationpred}

Traditional methods such as Nearest Class Mean (NCM) \cite{ma-etal-2024-making} predict relations by selecting the class whose prototype has the smallest distance to the test sample $x$. While effective, this approach relies solely on distance metrics, which may not fully capture the nuanced relationships between a sample and the broader context provided by class descriptions.

Rather than merely seeking the closest prototype, we aim to retrieve the class description that best aligns with the input, thereby leveraging the inherent semantic meaning of the label. To achieve this, we introduce \emph{Descriptive Retrieval Inference} (DRI), a retrieval mechanism fusing two distinct reciprocal ranking scores. This approach not only considers the proximity of a sample to class prototypes but also incorporates cosine similarity measures between the sample’s hidden representation $\bm{z}$ and relation descriptions generated by an LLM. This dual focus on both spatial and semantic alignment ensures that the final prediction is informed by a richer, more robust understanding of the relations.

Given a sample $x$ with hidden representation $\bm{z}$ and a set of relation prototypes $\{\bm{p}_r\}_{r=1}^n$, the inference process begins by calculating the negative Euclidean distance between $\bm{z}$ and each prototype $\bm{p}_r$:
\begin{align}
&\textbf{E}(x,r) = -{\left\| \bm{z} - \bm{p}_r \right\|}_2,\\
&\bm{p}_r = \frac{1}{L} \sum_{i=1}^{L}\bm{z}_i, 
\end{align}
where $L$ is the memory size per relation. Simultaneously, we compute the cosine similarity between the hidden representation and each relation description prototype, $\gamma(\bm{z}, \bm{d}_r)$. These two scores are combined into DRI score of sample $\bm{x}$ w.r.t relation $r$ for inference, ensuring that predictions align with both label prototypes and relation descriptions:
\begin{align}
\begin{aligned}
\textrm{DRI}(x, r) = &\frac{\alpha}{\epsilon + \text{rank}(\textbf{E}(x, r))} + \frac{1 - \alpha}{\epsilon + \text{rank}(\gamma(\bm{z}, \bm{d}_r))},
\end{aligned}
\end{align}
where $d_r = \frac{1}{K}\sum_{i=1}^{K} d_r^i$, $\text{rank}(\cdot)$ represents the rank position of the score among all relations. The $\alpha$ hyperparameter balances the contributions of the Euclidean distance-based score and the cosine similarity score in the final ranking for inference, and $\epsilon$ is a hyperparameter that controls the influence of lower-ranked relations in the final prediction. By adjusting $\epsilon$, we can fine-tune the model's sensitivity to less prominent relations. Finally, the predicted relation label $y^*$ is predicted as the one corresponding to the highest DRI score:
\begin{equation}
y_x^* = \underset{r=1, \ldots, n}{\operatorname{argmax}} \, \text{DRI}(x,r)
\end{equation}

This fusion approach for inference complements the learning paradigm, ensuring consistency and reliability throughout the FCRE process. By effectively balancing the strengths of protoype-based proximity and description-based semantic similarity, it leads to more accurate and robust predictions across sequential tasks.


\section{Experiments}


\subsection{Settings}
\label{sec:exp:setup}

 We conduct experiments using two pre-trained language models, BERT \citep{devlin-etal-2019-bert} and LLM2Vec \cite{llm2vec}, on two widely used benchmark datasets for Relation Extraction: FewRel \citep{han-etal-2018-fewrel} and TACRED \citep{zhang-etal-2017-position}. We benchmark our methods against state-of-the-art baselines: \textbf{SCKD} \cite{wang-etal-2023-serial},  \textbf{RP-CRE} \cite{cui-etal-2021-refining},  \textbf{CRL} \cite{zhao-etal-2022-consistent}, \textbf{CRECL} \cite{hu-etal-2022-improving}, \textbf{ERDA} \cite{qin-joty-2022-continual}, \textbf{ConPL} \cite{DBLP:conf/acl/ChenWS23}, \textbf{CPL} \cite{ma-etal-2024-making}, \textbf{CPL+MI} \cite{tran-etal-2024-preserving}.

\subsection{Experiment results}
 \label{exp_method}

\paragraph{Our proposed method yields state-of-the-art accuracy.}
Table \ref{table:main} presents the results of our method and the baselines, all using the same pre-trained BERT-based backbone. Our method consistently outperforms all baselines across the board. The performance gap between our method and the strongest baseline, CPL, reaches up to $3.74\%$ on FewRel and $5.82\%$ on TACRED.

To further validate our model, we tested it on LLM2Vec, which provides stronger representation learning than BERT. As shown in Table \ref{table:main_llmvec}, our model again surpasses CPL, with accuracy drops of only $16.58\%$ on FewRel and $9.96\%$ on TACRED.

These results highlight the effectiveness of our method in leveraging semantic information from descriptions, which helps mitigate forgetting and overfitting, ultimately leading to significant performance improvements.


\begin{figure}[ht]
	\centering
	\includegraphics[width=\linewidth]{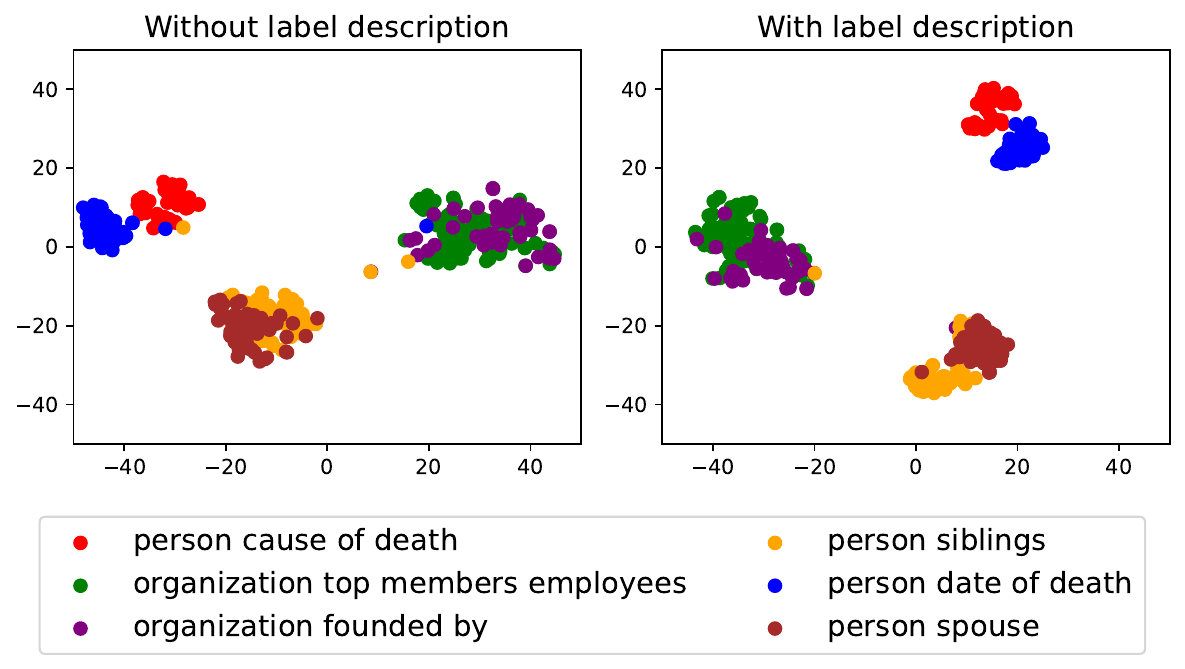}
	\caption{t-SNE visualization of the representations of 6 relations post-training, with and without descriptions, using our retrieval strategy.}
	\label{fig:tnse1}
\end{figure}

\begin{figure}[ht]
	\centering
	\includegraphics[width=\linewidth]{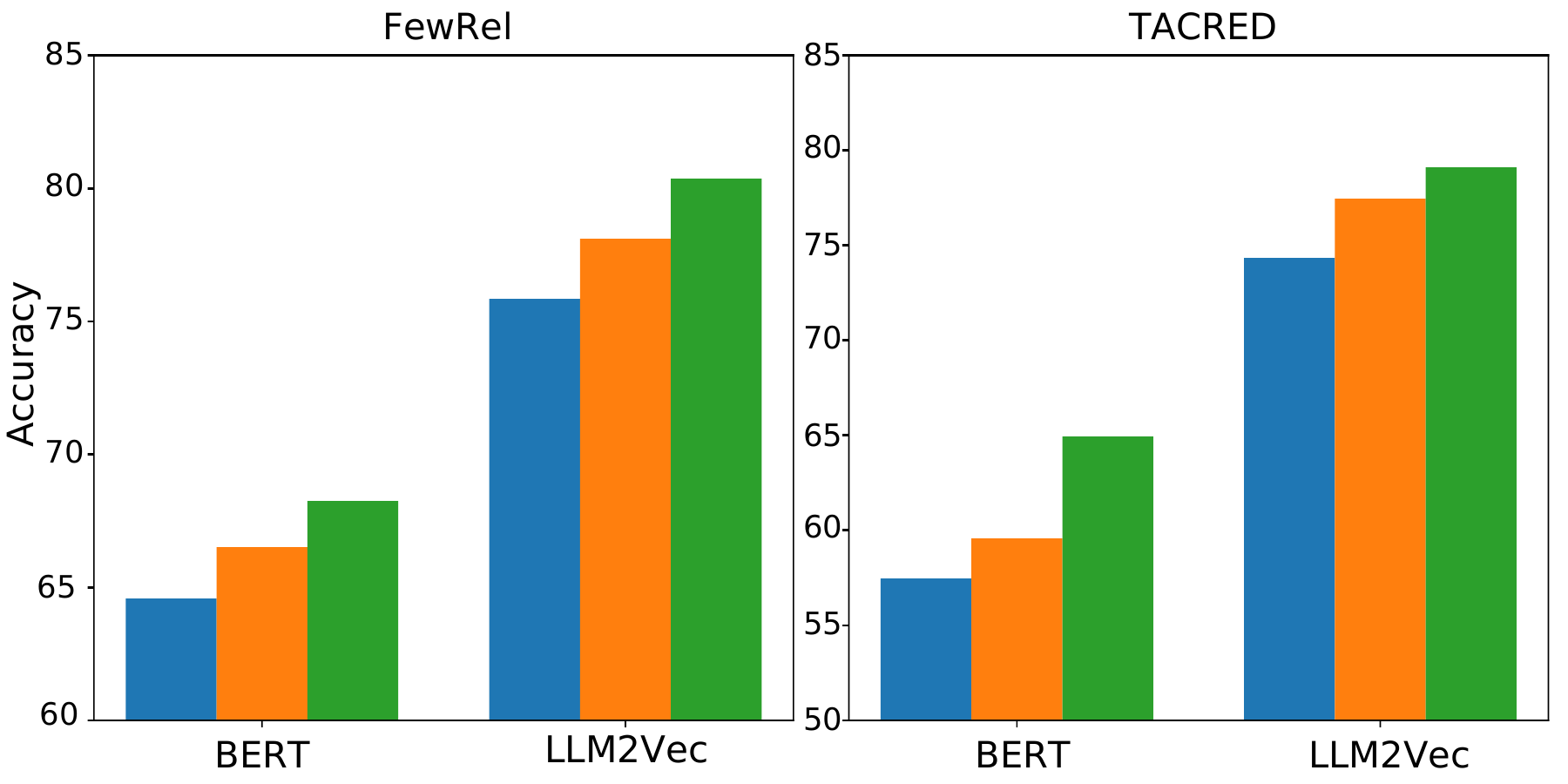}
	\caption{The impact of refined descriptions generated by LLMs. The green, orange, and blue bars show respectively the final accuracies of DCRE when using refined descriptions, original descriptions, and without using descriptions.}
	\label{fig:expert_or_not}
\end{figure}

\begin{figure}[h]
	\centering
	\includegraphics[width=\linewidth]{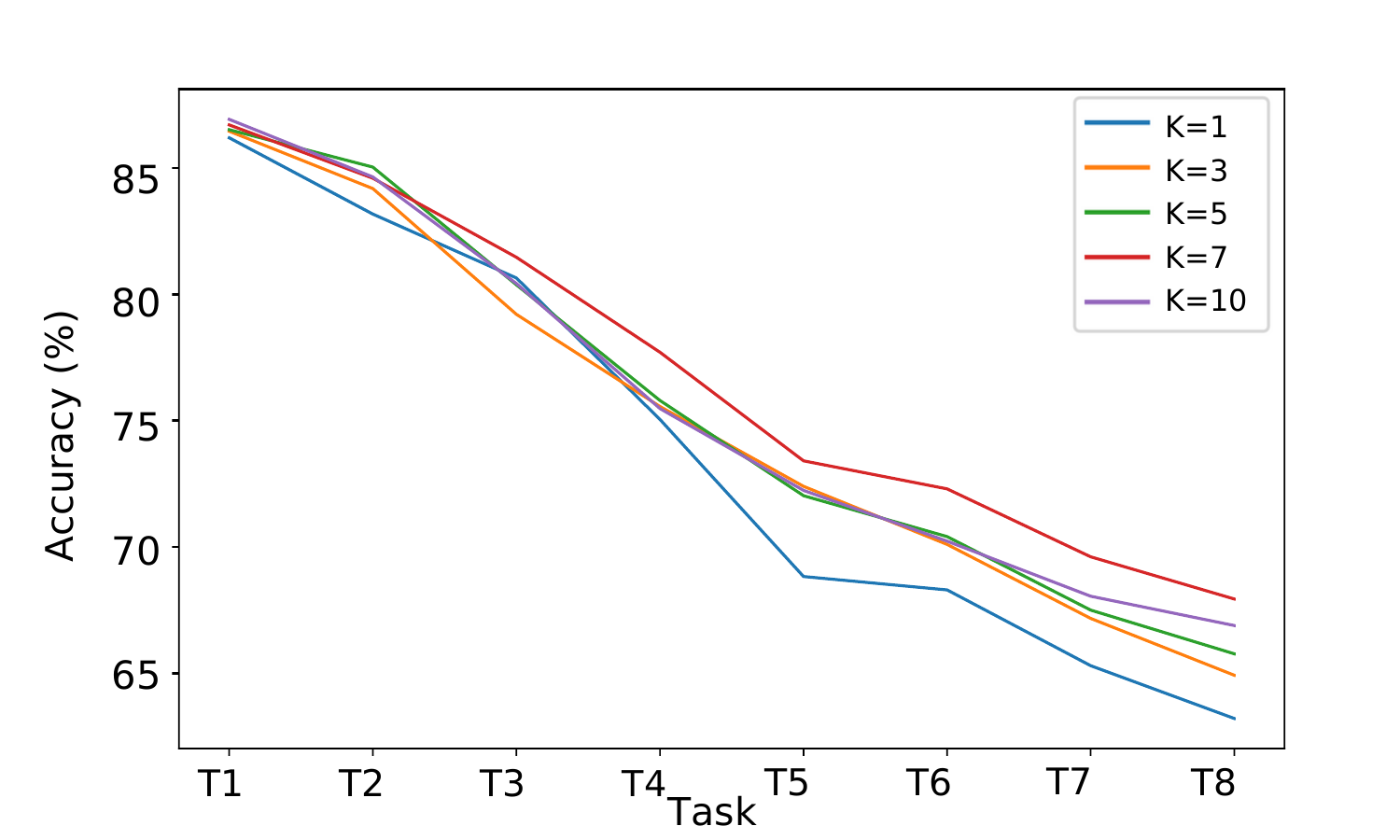}
	\caption{Model performance when varying K, on \textit{TACRED 5-way 5-shot}.}
	\label{fig:vary_k}
\end{figure}

\paragraph{Exploiting additional descriptions significantly enhances representation learning.}
Figure \ref{fig:tnse1} presents t-SNE visualizations of the latent space of relations without (left) and with (right) the use of descriptions during training. The visualizations reveal that incorporating descriptions markedly improves the quality of the model’s representation learning. For instance, the brown-orange and purple-green class pairs, which are closely clustered and prone to misclassification in the left image, are more distinctly separated in the right image. Additionally, Figure \ref{fig:expert_or_not} illustrates that our strategy, which leverages refined descriptions, captures more semantic knowledge related to the labels than the approach using raw descriptions. This advantage bridges the gap imposed by the challenges of few-shot continual learning scenarios, leading to superior performance. Figure \ref{fig:vary_k} shows the perfomance of our model on TACRED as the number of generated expert descriptions per training varies. The results indicate that the model performance generally improves from $K = 3$ and peaks at $K = 7$. 


\paragraph{Our retrieval-based prediction strategy notably enhances model performance.}
Table \ref{tab:prediction} demonstrates that by leveraging the rich information from generated descriptions, our proposed strategy improves the model’s performance by up to $1.31\%$ on FewRel and $6.66\%$ on TACRED compared to traditional NCM-based classification. The harmonious integration of NCM-based prototype proximity and description-based semantic similarity enables our strategy to deliver more accurate and robust predictions across sequential tasks.

\begin{table}[!t]
    \centering
    \setlength{\tabcolsep}{1mm}
    \begin{tabular}{lcccc}
    \hline 
    \multirow{2}{*}{Method} & \multicolumn{2}{c}{FewRel} & \multicolumn{2}{c}{TACRED}\\
    \cmidrule(lr){2-3} \cmidrule(lr){4-5} 
    & BERT & LLM2Vec & BERT & LLM2Vec  \\
    \hline \hline \\
    NCM  & 66.93 & 79.26 & 58.26 & 75.00 \\ 
    DRI (Ours) &\textbf{ 68.24} & \textbf{80.34} & \textbf{63.21} & \textbf{79.10}\\
    \hline
    \end{tabular}
    \caption{DRI and NCM prediction.}
    \label{tab:prediction}
\end{table}



\subsection{Ablation study}
\label{sec:exp:ablation}

Table \ref{tab:ablation_study} present evaluation results that closely examine the role of each component in the objective function during training. The findings underscore the critical importance of $\mathcal{L}_\text{MI}$ and $\mathcal{L}_\text{HM}$, both of which leverage instructive descriptions from LLMs, aided by \emph{Raw descriptions}. Because when we ablate one of them, the final accuracy can be reduced by 6\% on the BERT-based model, and 10\% on the LLM2VEC-based model.

\begin{table}[ht]
    \centering
    \setlength{\tabcolsep}{1mm}
    \begin{tabular}{lcccc}
    \hline 
    \multirow{2}{*}{Method} & \multicolumn{2}{c}{BERT} & \multicolumn{2}{c}{LLM2Vec}\\
    \cmidrule(lr){2-3} \cmidrule(lr){4-5} 
    & FewRel & TACRED & FewRel & TACRED \\
    \hline \hline \\
    \text{DCRE (Our)} & \textbf{68.24} & \textbf{63.21} & \textbf{80.34} & \textbf{79.10}\\
    \quad  \small \text{w/o $\mathcal{L}_{\textrm{SC}}$} & \underline{67.58} &  62.11 & \underline{78.39} & \underline{77.01}\\
    \quad  \small \text{w/o $\mathcal{L}_{\textrm{MI}}$} & 65.10 & 57.23 & 70.61 & 74.17\\
    \quad  \small \text{w/o $\mathcal{L}_{\textrm{HM}}$} & 66.20 & \underline{62.46} & 77.22 & 74.75\\
    \quad  \small \text{w/o $\mathcal{L}_{\textrm{ST}}$} & 67.54 & 59.56 & 77.48 & 73.77\\
    \hline
    \end{tabular}
    \caption{Ablation study.}
    \label{tab:ablation_study}
\end{table}

\section{Conclusion}
In this work, we propose a novel retrieval-based approach to address the challenging problem of Few-shot Continual Relation Extraction. By leveraging large language models to generate rich relation descriptions, our bi-encoder training paradigm enhances both sample and class representations and also enables a robust retrieval-based prediction method that maintains performance across sequential tasks. Extensive experiments demonstrate the effectiveness of our approach in advancing the state-of-the-art and overcoming the limitations of traditional memory-based techniques, underscoring the potential of language models and retrieval techniques for dynamic real-world relationship identification.

\bibliography{aaai25, anthology}


%

\end{document}